\documentclass[letterpaper, 10 pt, conference]{ieeeconf}  

\IEEEoverridecommandlockouts                              
\usepackage{amsmath}
\usepackage{booktabs}
\usepackage[pdftex]{graphicx}
\usepackage{hyperref}
\hypersetup{colorlinks=true, citecolor=black, linkcolor=black}
\hypersetup{urlcolor=black}
\hypersetup{bookmarksnumbered=true}
\hypersetup{plainpages=false}
\usepackage[all]{hypcap} 
\usepackage{subfig}

\usepackage{textcomp}

\title{\LARGE \bf
Stochastic Collection and Replenishment (SCAR) Optimisation for Persistent Autonomy
}

\author{Andrew W Palmer, Andrew J Hill and Steven J Scheding$^{1}$
\thanks{This work has been supported by the Rio Tinto Centre for Mine Automation and the Australian Centre for Field Robotics, University of Sydney, Australia.}%
\thanks{$^{1}$The authors are with the Australian Centre for Field Robotics, University of Sydney, Australia. Email addresses: {\tt\small \{a.palmer;a.hill;s.scheding\}@acfr.usyd.edu.au}}%
\thanks{\textcopyright 2014 IEEE. Personal use of this material is permitted. Permission from IEEE must be obtained for all other uses, in any current or future media, including reprinting/republishing this material for advertising or promotional purposes, creating new collective works, for resale or redistribution to servers or lists, or reuse of any copyrighted component of this work in other works.}
}

\begin{document}
\maketitle
\thispagestyle{empty}
\pagestyle{empty}

\begin{abstract}

Robots have a finite supply of resources such as fuel, battery charge, and storage space. The aim of the Stochastic Collection and Replenishment (SCAR) scenario is to use dedicated agents to refuel, recharge, or otherwise replenish robots in the field to facilitate persistent autonomy. This paper explores the optimisation of the SCAR scenario with a single replenishment agent, using several different objective functions. The problem is framed as a combinatorial optimisation problem, and A* is used to find the optimal schedule. Through a computational study, a ratio objective function is shown to have superior performance compared with a total weighted tardiness objective function, with a greater performance advantage present when using shorter schedule lengths. The importance of incorporating uncertainty in the objective function used in the optimisation process is also highlighted, in particular for scenarios in which the replenishment agent is under- or fully-utilised. 

\end{abstract}

\section{INTRODUCTION}

Persistent autonomy requires consideration of resources such as battery charge and data storage space. In many cases the solution to the energy replenishment problem is to either return to a charging dock \cite{Luo2005} or collect energy from the environment \cite{Nguyen2013}. Similarly, wireless technologies can remove the limitation of data storage space on the operation time of an agent. However, there are many cases where these methods are sub-optimal or even impossible. Surveillance tasks, for example, require the agent to remain in the surveillance area \cite{Mathew2013}, and wireless communications are generally troublesome in underwater environments \cite{Vasilescu2005}. One method of addressing these issues is to use another agent to travel to the agents in the field to recharge or transfer data with them. 

The Stochastic Collection and Replenishment (SCAR) scenario focuses on the use of dedicated replenishment agents to replenish the resource supply of the user agents working out in the field, thus enabling the user agents to remain active in the field indefinitely. Examples of possible resources include fuel, battery charge, food, and water for the replenishment case, and electronic data and physical samples for the collection case. The SCAR scenario was introduced in the authors' previous work to address several shortcomings in the literature \cite{Palmer2013}. This paper builds on the previous work by performing schedule optimisation in the SCAR scenario, and examining the performance of the ratio objective function introduced in \cite{Palmer2013} in comparison to the total tardiness objective function. The main contribution of this paper is the implementation of a schedule optimisation for the SCAR scenario with a single replenishment agent. In doing so, an evaluation of the ratio objective function is presented, along with an admissible and consistent heuristic for using the ratio objective function in graph search methods. The benefit of incorporating uncertainty when calculating the objective function is also demonstrated. 

The rest of this paper is structured as follows: Section \ref{s:probdef} provides an overview of the assumptions and characteristics of the SCAR scenario, and Section \ref{s:relatedlit} covers the related literature. A discussion on schedule optimisation in the SCAR scenario is given in Section \ref{s:opt}, with the performance of several objective functions evaluated in a computational study in Section \ref{s:compstudy}. Finally, some concluding remarks are given in Section \ref{s:conc}. 

\section{PROBLEM DEFINITION} \label{s:probdef}

The SCAR scenario is a generalised collection or replenishment scenario that was first defined in \cite{Palmer2013}. The replenishment scenario consists of multiple user agents that consume a resource over time. They have a limited capacity of the resource and are unable to continue operating when their supply of the resource is depleted. To enable persistence, replenishment agents will periodically rendezvous with the user agents and transfer a quantity of the resource to them to replenish their resource supply. The collection scenario is mathematically identical to the replenishment scenario except for the direction of the resource flow. To replenish a user agent, the replenishment agent must travel to the user agent, set-up the connection between the agents before beginning replenishment, replenish the user agent and then pack-up before travelling to the next user agent. The user agent may continue to use the resource during replenishment. The replenishment agents also have a limited capacity of the resource, and they must periodically return to a replenishment point to replenish their supply of the resource. 

Both the fleet of user agents and the fleet of replenishment agents are heterogeneous. Parameters such as the speed of the agents, resource usage and replenishment rates, and set-up and pack-up times are all stochastic in nature. The user agents may be positioned apart from one another, and the motion of the agents may be constrained by roads or obstacles. A replenishment agent can service only one user agent at a time, and similarly each user agent can only be serviced by one replenishment agent at a time. There are no limits on how many times a user or replenishment agent may be replenished. Once replenishment has started, it must continue until either the user agent is full or the replenishment agent has exhausted its resource supply. 

The consequences of a user agent exhausting its resource supply depend on the scenario under investigation. For example, if an Unmanned Aerial Vehicle (UAV) runs out of fuel mid-flight, it will crash resulting in the loss of the agent. This represents a hard deadline and can be modelled by a step cost. A soft deadline on the other hand relies on the agent entering a safe zero-energy state. For example, an Unmanned Ground Vehicle (UGV) that runs out of fuel, comes to a safe stop, and is able to resume operation once replenished. However, this depends on the environment that the agent is operating in---stopping in the middle of a highway would not be considered safe. A soft deadline results in a cost that increases with the time that the agent is empty. 

A scenario with a single replenishment agent and soft deadlines has been assumed for this paper. It has also been assumed that the user agents operate in their own small operational areas (eg. they are performing a local surveillance task). Travel time variations due to the movements of the user agents are then accounted for in the set-up and pack-up times of the replenishment agent. The aim when optimising this scenario is to determine a schedule for the replenishment agent which will minimise the time that the user agents are without the resource and therefore not operational. 

\section{RELATED LITERATURE} \label{s:relatedlit}

The SCAR scenario combines the characteristics of a number of different replenishment and collection scenarios present in the literature. The main replenishment scenarios in the literature focus on the refuelling of aerial and space vehicles. Jin et al. \cite{Jin2006a}, Barnes et al. \cite{Barnes2004}, and Kaplan and Rabadi \cite{Kaplan2012}, \cite{Kaplan2013} examined the refuelling of aerial vehicles using dedicated tanker aircraft. They framed the problem as an NP-hard combinatorial optimisation problem and used heuristics such as the Apparent Tardiness Cost (ATC), meta-heuristics such as Simulated Annealing, and exact methods such as Dynamic Programming, to calculate schedules for the replenishment aircraft. Jin et al. \cite{Jin2006a} considered the refuelling of multiple UAVs by a single tanker aircraft and noted that it resembles a restricted Travelling Salesman Problem (TSP) with time windows. Kaplan and Rabadi \cite{Kaplan2012}, \cite{Kaplan2013} modelled the multiple tanker problem as a manufacturing job shop with parallel machines and applied classical scheduling methods to the scenario. 

Barnes et al. \cite{Barnes2004} considered the same situation as Kaplan and Rabadi, but with the inclusion of travel times between aircraft. They incorporated the travel times into the set-up times for each refuelling task, and modelled the scenario as a parallel machine manufacturing job shop with sequence dependent set-up times. The refuelling of satellites via a dedicated refuelling space vehicle was examined by Shen and Tsiotras \cite{Shen2002}, and is similar to the scenario considered by Barnes et al. in that the travel between the satellites is included in the cost calculation. The scenario presented in Mathew et al. \cite{Mathew2013} involved the recharging of multiple UAVs performing a surveillance task. Instead of using an aerial tanker, a UGV was used to rendezvous with and recharge the UAVs. 

The main collection scenario in the literature involves the use of data mules to collect data from a remote sensor network. A motivating example for this given in Vasilescu et al. \cite{Vasilescu2005} is the long-term monitoring of coral reefs and other underwater environments. As long range underwater communication is difficult, they proposed the use of an Autonomous Underwater Vehicle (AUV) that travels to each sensor and gathers the data that the sensor has collected. Similar collection scenarios are presented in \cite{Dunbabin2006}, \cite{Yuan2007}, \cite{Bhadauria2011} and \cite{Tekdas2012}. In most cases, the authors treat the problem as a variant of the TSP and solve accordingly. 

The above literature has a number of limiting assumptions. First of all, the replenishment agents are assumed to have sufficient supply of the resource to fully replenish all of the user agents. However, over the long term, this will not be the case and the replenishment agent will have to return to the replenishment point to replenish its own supply of the resource. Secondly, the user agents are visited only once. Again, over the long term, the user agents will have to be replenished more than once. This is a particularly important consideration in cases where some user agents exhaust their supply of the resource significantly faster than others. Thirdly, the time to replenish the user agents in most cases is assumed to be constant. Depending on the scenario, the effect of replenishment times that vary over time may be quite significant. Finally, all of the above literature treats the scenario as deterministic. For real-world applications this is not realistic as the speeds, usage rates, and other agent parameters are going to have some element of uncertainty. The SCAR scenario addresses the above limitations by considering a limited capacity for the replenishment agent, multiple replenishments of user agents, variable replenishment times, and uncertainty in the agent parameters. 

Uncertainty has been examined in path planning problems \cite{Murphy2013}, \cite{Pereira2013}, but stochastic scheduling problems have received limited study due to their difficulty in comparison to deterministic problems \cite{Allahverdi2008}. A common way of dealing with uncertainty is to replan when the state of the system changes \cite{Ferguson2005}. Typically this happens in response to unexpected events or new information, such as finding new obstacles during exploration. Other work specifies a time period after which replanning is performed if no unexpected events have occurred. This allows short term deviations from the original plan or schedule to be captured in the optimisation process when considering a long term goal \cite{Munirathinam1994}. However, these methods do not take into account the risks associated with each task. Some methods of minimising risk when scheduling include using chance constraints \cite{Orcun1996}, conservative estimates of the uncertain parameters \cite{Bertsimas2003}, or estimation of an expected cost through Monte Carlo simulation \cite{Bassett1997}. 

In the authors' previous work, a mathematical framework for the SCAR scenario with a single replenishment agent was introduced \cite{Palmer2013}. Both a Monte Carlo method using values sampled from the parameter probability distributions, and an analytical method which used the entire parameter probability distributions, were introduced for calculating the expected cost of a schedule. The analytical method was shown to produce comparable results to the Monte Carlo method while computing significantly faster. 

\section{OPTIMISATION} \label{s:opt}

A schedule is a list of the tasks that the replenishment agent is to perform in the order of execution. The two possible tasks are to replenish one of the user agents, or for the replenishment agent to travel to the replenishment point to be replenished itself. The task of replenishing a user agent is denoted by the index of the user agent to be replenished, while returning to the replenishment point is denoted by $r$. An example schedule of five tasks where the replenishment agent is firstly replenished at the replenishment point, followed by replenishing 4 user agents in index order would be represented by: 

\[(r, 0, 1, 2, 3)\]

It is important to note that the assumptions of the SCAR scenario allow for user agents to be replenished multiple times, so tasks can appear multiple time within a schedule. However, there is no requirement for a particular task to appear in the schedule. Therefore, it is possible to end up with a schedule such as:
\[(0,r,3,2,0,2,r)\]
where the task of replenishing user agent 1 does not appear. 

The set of all possible schedules can be represented as a tree, where each node is a task that branches to all of the possible tasks at the next layer. The root of the tree is a phantom node that does not represent any task. Tasks appear multiple times in each layer, but have different parent nodes. 

\subsection{Objective Functions} \label{s:objfunc}

The literature presented in Section \ref{s:relatedlit} use a number of different objective functions in the optimisation. Examples include sum of priority weighted time of refuelling \cite{Jin2006a}, total weighted tardiness \cite{Kaplan2012}, travel cost \cite{Mathew2013}, travel distance \cite{Yuan2007}, and total time \cite{Tekdas2012}, while a hierarchical objective function is used in \cite{Barnes2004} to optimise over multiple objectives. In the SCAR scenario, the aim is to minimise the total time that the user agents are expected to be without the resource. This best corresponds with the total weighted tardiness used in \cite{Kaplan2012}. The tardiness, $c_{x}$, of a task, $x$, is defined as:
\[c_{x} = \max(0, f_{x}-d_{x})\]
where $f_{x}$ is the completion time and $d_{x}$ is the deadline of the task. 

For the SCAR scenario, the deadline refers to the time before which replenishment must start to prevent the user agent from exhausting its supply of the resource, and the completion time is instead the time that the replenishment begins. The expected total weighted tardiness, $E[C]$, in the SCAR scenario corresponds to the total time that the user agents' resource supplies are expected to be empty, weighted by their priorities. It is formally defined as:

\begin{equation}
\arg \min E[C] = \sum_{i=0}^{n-1} w_{i} E[T_{i}]
\end{equation}
where $E[T_{i}]$ is the expected time that user agent $i$ is empty during the schedule, $w_{i}$ is the weighting applied to user agent $i$, and $n$ is the number of user agents in the system. 

The expected ratio cost, $E[R]$, is a modified version of the expected total weighted tardiness measure which normalises the expected total weighted tardiness measure by the expected total schedule time and the number of user agents. It is defined as:

\begin{equation} \label{eq:ratio}
\arg \min E[R] = \frac{\displaystyle\sum_{i=0}^{n-1} w_{i} E[T_{i}]}{n E[T_{max}]}
\end{equation}
where $E[T_{max}]$ is the total time of the schedule. Provided
\[\sum_{i=0}^{n-1} w_{i} = 1\]
then
\[0 \le E[R] \le 1 \]

If $w_{0} = w_{1} = ... = w_{n}$, the expected ratio cost is equivalent to the proportion of time during the schedule that the user agents are expected to be empty. 

The main benefit of using the ratio cost is greater comparability between schedules, as any two given schedules are unlikely to take exactly the same amount of time to be executed by the replenishment agent. As an example, consider two schedules, $A$ and $B$, for a scenario consisting of 4 user agents. The total weighted tardiness, total schedule time, and ratio cost for each schedule are shown in Table \ref{t:example}. As can be seen, while schedule $A$ has the lower total weighted tardiness, in schedule $B$ the user agents are actually empty for a lower proportion of time. 

\begin{table}[b]
\caption{Expected Ratio Cost Example}
\label{t:example}
\centering
\begin{tabular}{ c  c  c}
\toprule
  Schedule & A & B  \\
  \midrule
  Tardiness & 500 & 520  \\
  Total time & 1000 & 1100 \\
  \midrule
  Ratio & 0.125 & 0.118  \\
  \bottomrule
\end{tabular}
\end{table}

In this paper, four objective functions are tested:

\begin{itemize}
\item Deterministic Total Weighted Tardiness (DT)
\item Stochastic Total Weighted Tardiness (ST)
\item Deterministic Ratio (DR)
\item Stochastic Ratio (SR)
\end{itemize}

The objective functions will be referred to by their acronyms for the remainder of this paper. The deterministic forms of the objective functions use the mean values of the parameters, such as speed and resource usage rate, to calculate the expected value, while ignoring the uncertainty associated with those parameters. The stochastic forms, on the other hand, use the analytical method presented in the authors' previous work \cite{Palmer2013} to generate an expected value which takes into consideration the parameter uncertainty. The stochastic forms of the objective functions allow risk-based discrimination between schedules that would produce similar costs when using the deterministic forms of the objective functions.

\subsection{Applying A* to SCAR Optimisation} \label{s:astar}

A* is just one of many graph search algorithms that could be used to optimise this scenario, and is only being used to illustrate the importance of the choice of objective function. In path planning, A* is used to find a path from a starting node to a goal node. However, for scheduling, there is no goal task that the schedule must finish on. Instead, the aim is to find the lowest cost schedule. Given the persistent nature of the tasks being performed by the user agents, a full schedule for the SCAR scenario consists of an infinite number of tasks. In order to perform optimisation on the schedule of the replenishment agent, a finite horizon has been used. The finite horizon used in this paper consists of a set number of tasks. A phantom node is added at a depth of $h+1$ in the tree, where $h$ is the number of tasks in the finite horizon. The first solution to reach the phantom node is the lowest cost schedule. 

Two limitations are placed on the schedules to avoid searching through schedules that produce undesirable behaviour. Firstly, it doesn't make sense to perform a task twice in a row, so a node in the tree is not allowed to branch to a node of the same task. Secondly, if the replenishment agent has exhausted its supply of the resource, or the resource level is below a threshold, the only allowed task is to send it to the replenishment point. A threshold of 5\% was found to give good results in the scenario presented in this paper. In general, this figure should be selected to be less than the capacity of a user agent to ensure that the replenishment agent is not sent back to the replenishment point too early. 

When using A*, it is important to select a heuristic that is both admissible and consistent to ensure that the returned solution is optimal. An admissible heuristic never overestimates the cost from the current node to the goal, and a consistent heuristic is monotonic \cite{Murphy2013}. Creating a good heuristic which is admissible and consistent for the SCAR scenario is challenging. For the DT and ST objective functions, the most simple but na\"{i}ve heuristic of assuming that no user agents are empty to the end of the schedule has been used. This essentially reduces the A* search to a best-first search as the cost at the current node is the cost estimate for the total schedule. 

For the DR and SR objective functions, $E[T_{max}]$ from (\ref{eq:ratio}) must be estimated. It is important not to underestimate $E[T_{max}]$ as this can lead to the heuristic being inadmissible. A maximum value for $E[T_{max}]$ can be estimated by computing the longest time each task can take when following on from a previous task. By assuming that each user agent is completely empty, the maximum time for each individual task can be calculated. 

Calculating $E[T_{max}]$ cannot be performed in a greedy manner as this can underestimate the maximum schedule time. Instead, combinatorial optimisation is used to calculate the maximum time. While a method like A* could be used to calculate the maximum time in each instance, it is more efficient to use dynamic programming to precompute a lookup table of the maximum times to complete a schedule based on the previous task and the number of tasks to be selected. A standard backwards recursion dynamic programming approach similar to that used by Jin et al. \cite{Jin2006a} has been used in this paper for precomputing these maximum times.


The A* algorithm used in this paper replans after each task has been completed. This allows the differences between the actual and expected states to be incorporated in the optimisation process. This replanning occurs when using both the deterministic and stochastic forms of the objective functions. 

\section{COMPUTATIONAL STUDY} \label{s:compstudy}

\begin{figure}
  \centering
    \includegraphics[width=0.48\textwidth]{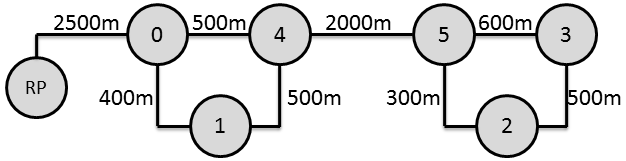}
  \caption{Layout showing the location of the Replenishment Point (RP) and User Agents}
  \label{f:layout}
\end{figure}

A computational study was performed on a scenario involving multiple UGVs performing a persistent task. The UGVs have a finite capacity of fuel which is replenished by a dedicated replenishment UGV. If they exhaust their supply of fuel, they enter a safe state. The replenishment agent has an independent supply of fuel, and it is assumed to be replenished when it visits the replenishment point. All of the agents are constrained to roads as shown in Fig.~\ref{f:layout}. The set of user agents is heterogeneous and their parameters are shown in Table \ref{t:param_u}. The replenishment agent parameters are shown in Table \ref{t:param_r}, and the parameters of the replenishment point are shown in Table \ref{t:param_p}. The user agents have equal priorities, and all uncertain parameters are assumed to be distributed according to Gaussian distributions.

\begin{table}
\caption{User Agent Parameters}
\label{t:param_u}
\centering
\begin{tabular}{ c  c  c  c  c  c  c}
\toprule
  Agent & 0 & 1 & 2 & 3 & 4 & 5 \\
  \midrule
  Capacity (L) & 1000 & 1200 & 700 & 1200 & 1000 & 800 \\
  Usage rate (L/s) & 0.5 & 0.5 & 0.3 & 0.5 & 0.4 & 0.4 \\
  mean & & & & & &  \\
  Usage rate (L/s) & 0.05 & 0.05 & 0.05 & 0.02 & 0.08 & 0.04 \\
   standard deviation & & & & & & \\
  \bottomrule
\end{tabular}
\end{table}

\begin{table}
\caption{Replenishment Agent Parameters}
\label{t:param_r}
\centering
\begin{tabular}{ c  c  c }
\toprule
Parameter & Mean & Standard Deviation \\
\midrule
Capacity (L) & 5000 & 0 \\
Replenishment rate (L/s) & 10 & 0.5 \\
Set-up time (s) & 60 & 20 \\
Pack-up time (s) & 20 & 5 \\
Speed (m/s) & 15 & 0.5 \\
\bottomrule
\end{tabular}
\end{table}

\begin{table}
\caption{Replenishment Point Parameters}
\label{t:param_p}
\centering
\begin{tabular}{ c  c  c }
\toprule
  Parameter & Mean & Standard Deviation \\
  \midrule
  Set-up time (s) & 30 & 10 \\ 
  Pack-up time (s) & 10 & 1 \\ 
  Replenishment rate (L/s) & 20 & 1 \\ 
  \bottomrule
\end{tabular}
\end{table}


The scenario was run using 4-, 5- and 6-user agents, representing scenarios where the replenishment agent is under-utilised, fully-utilised, and over-utilised respectively. The four objective functions introduced in Section \ref{s:objfunc} where tested individually for each case using a scheduling horizon based on the number of user agents in the system. Each simulation lasted for 5 hours of simulated time, and was repeated 40 times for each objective function to account for the variability due to the stochastic nature of the simulations. Starting resource levels for the replenishment agent and each user agent were randomly initialised between 50\% and 100\% of their maximum resource level to simulate realistic, in-progress conditions for each simulation. The replenishment agent was initially positioned at the replenishment point. 

\begin{figure}
  \centering
  \subfloat[4 user agents]{
    \includegraphics[width=0.47\textwidth]{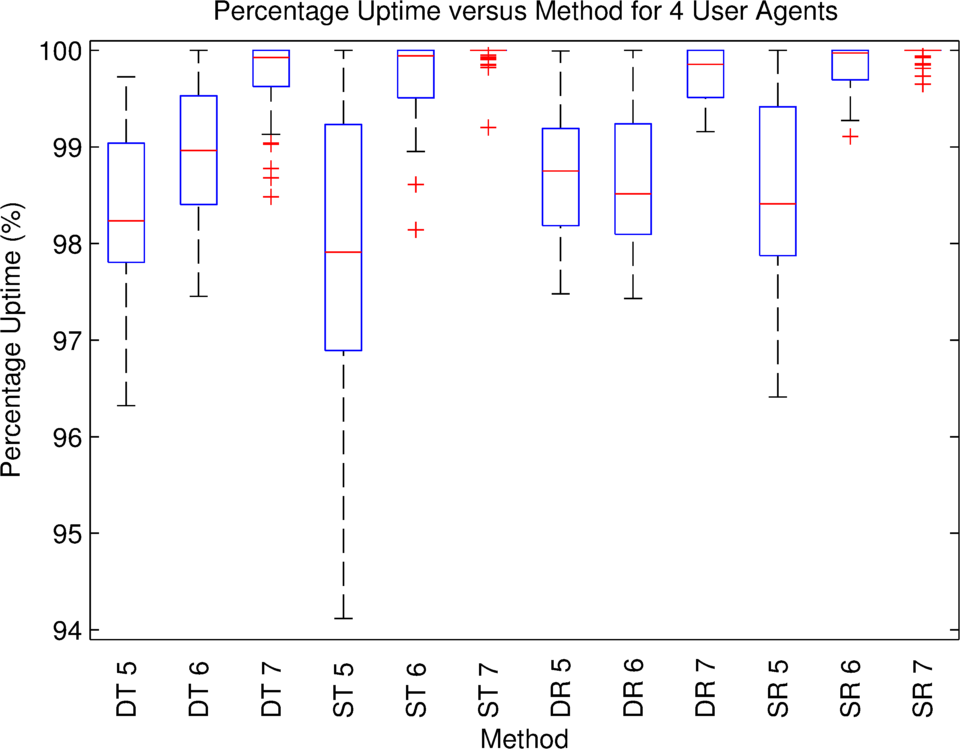}\label{sf:4_agent}
  }
  
  \subfloat[5 user agents]{
    \includegraphics[width=0.47\textwidth]{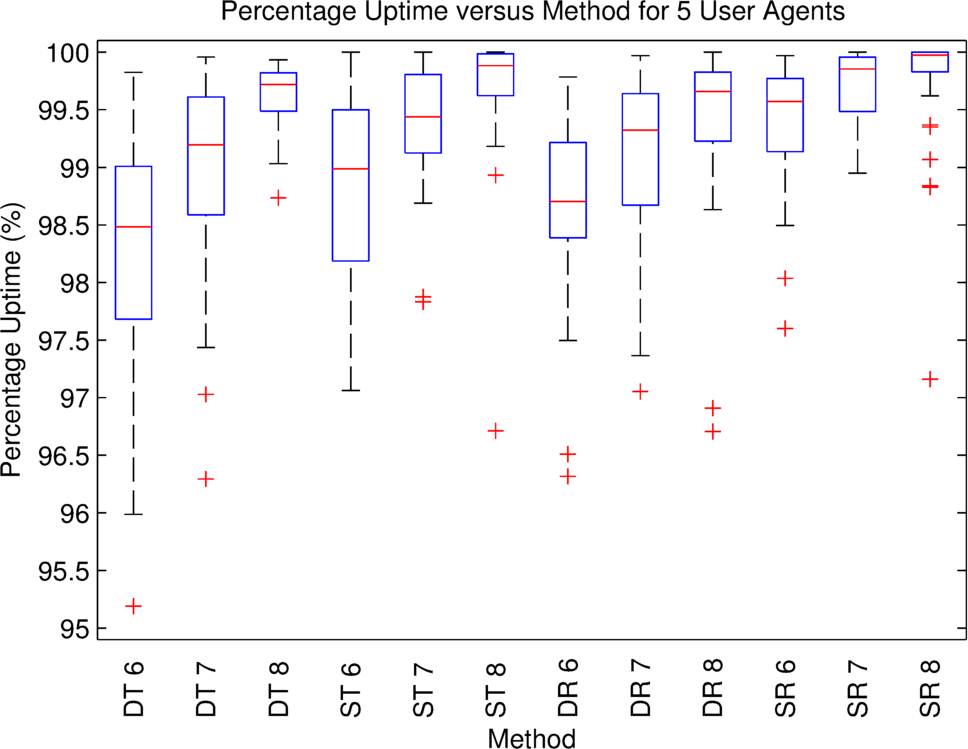}\label{sf:5_agent}
  }
  
  \subfloat[6 user agents]{
    \includegraphics[width=0.47\textwidth]{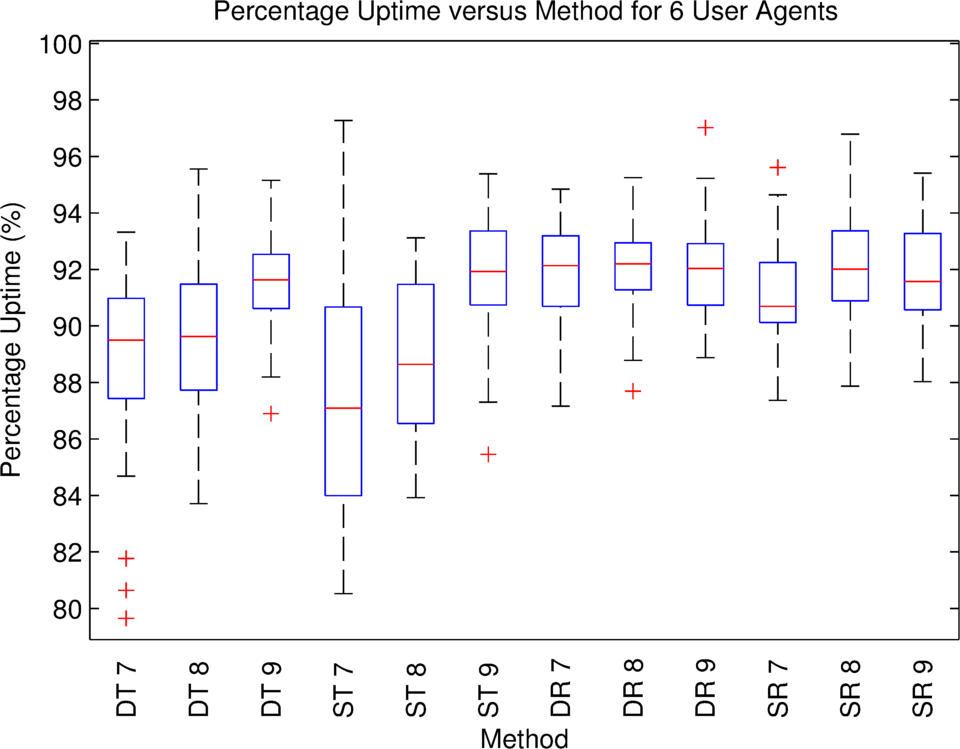}\label{sf:6_agent}
  }
  \caption{Box and whisker plots for the percentage uptime. The number next to the method name is the number of tasks in the schedule horizon.}
  \label{f:scen1}
\end{figure}

\begin{figure}
  \centering
  \subfloat[4 User Agents, Schedule Length of 5 Tasks]{
    \includegraphics[width=0.48\textwidth]{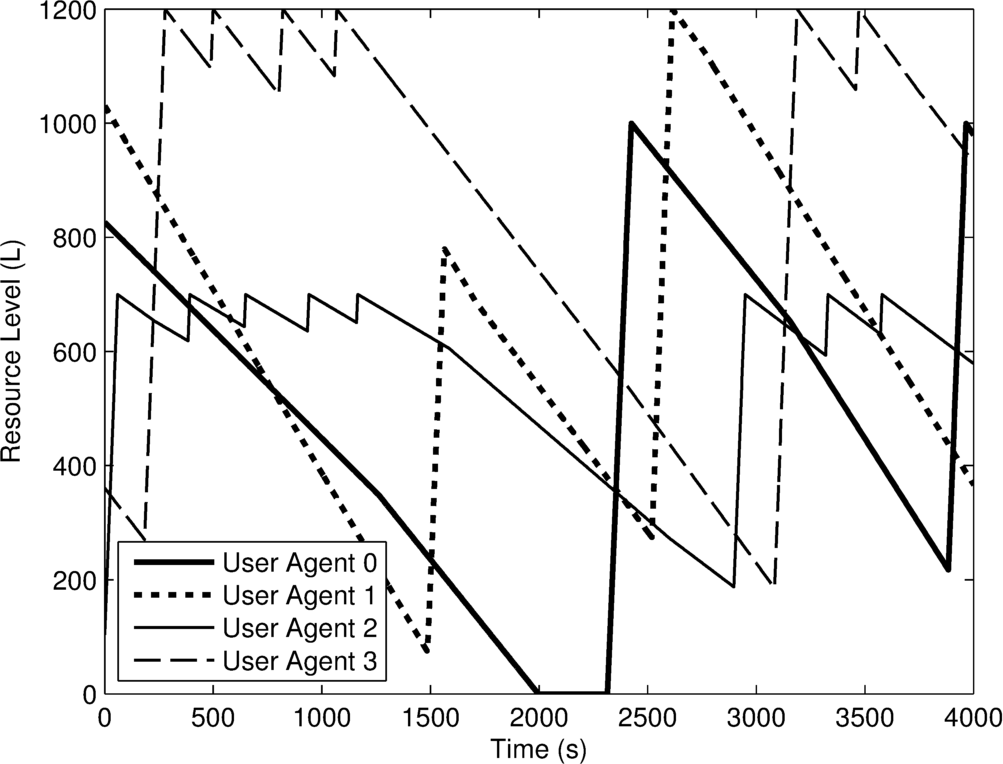}\label{sf:oscillations}
  }
  
 \subfloat[4 User Agents, Schedule Length of 12 Tasks]{
    \includegraphics[width=0.48\textwidth]{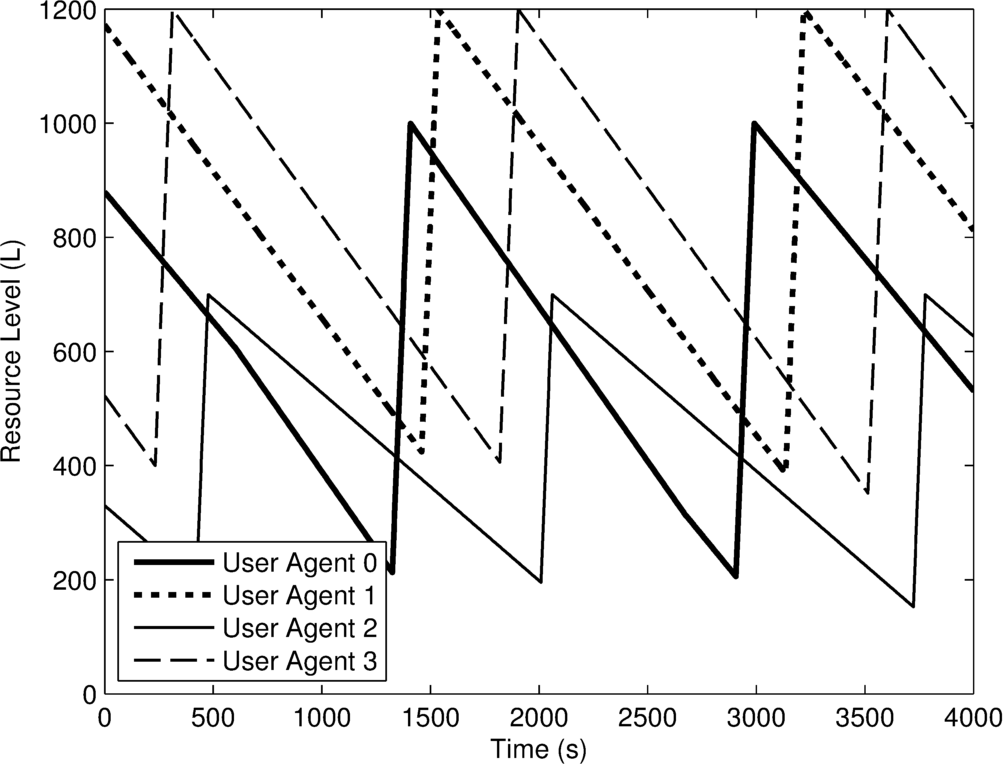}\label{sf:nooscillations}
  }
  \caption{Partial simulation results from the 4 user agent scenario, showing the user agent resource levels over time. Oscillations between short tasks in (a), as demonstrated by the saw-tooth pattern in agents 2 and 3 between 300s and 1200s, are a result of using a short schedule horizon. Using a longer schedule horizon in (b) minimises these oscillations. }
  \label{f:oscillations}
\end{figure}

The percentage uptime results for the 4-user agent scenario are shown in Fig.~\ref{sf:4_agent}. The DT and DR objective functions produced similar results, as did the ST and SR objective functions. The performance of all of the objective functions increased as the number of tasks in the schedule horizon was increased. When using short horizons, the objective functions favour short tasks, and this is demonstrated in Fig.~\ref{f:oscillations}. Fig.~\ref{sf:oscillations} shows that when using a schedule horizon of 5 tasks, there are clear oscillations between short tasks. Using a longer schedule horizon minimises these oscillations, with a schedule horizon of 12 tasks, shown in Fig.~\ref{sf:nooscillations}, completely removing the oscillations.  

\begin{figure}
  \centering
  \subfloat[4 user agents]{
    \includegraphics[width=0.47\textwidth]{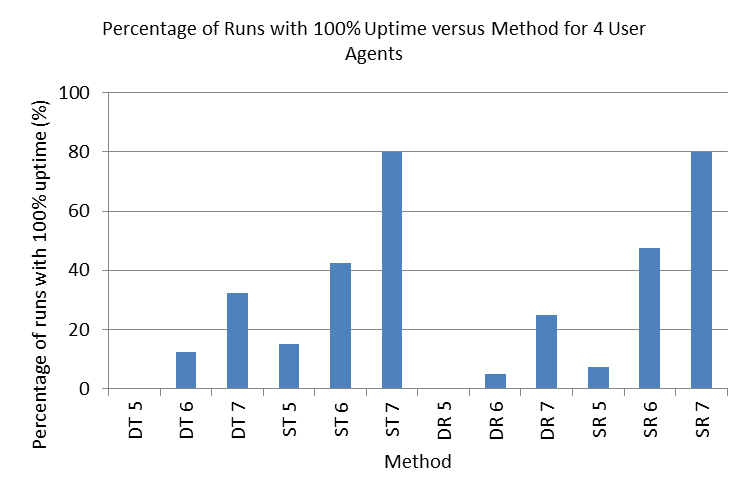}\label{sf:4_agent_2}
  }
  
  \subfloat[5 user agents]{
    \includegraphics[width=0.47\textwidth]{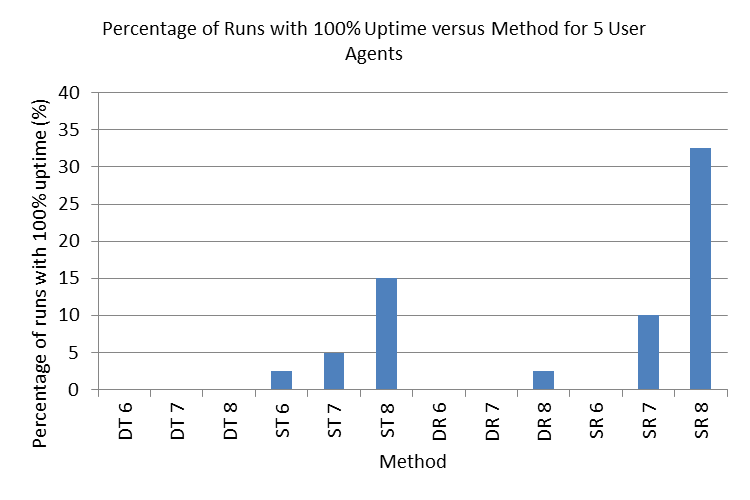}\label{sf:5_agent_2}
  }
..
  \caption{Percentage of simulation runs with 100\% uptime. The number next to the method name is the number of tasks in the schedule horizon. The 6 user agent scenario results are omitted as no method was able to achieve 100\% uptime in any of the simulations. }
  \label{f:scen1_2}
\end{figure}

\begin{figure}
  \centering
    \includegraphics[width=0.47\textwidth]{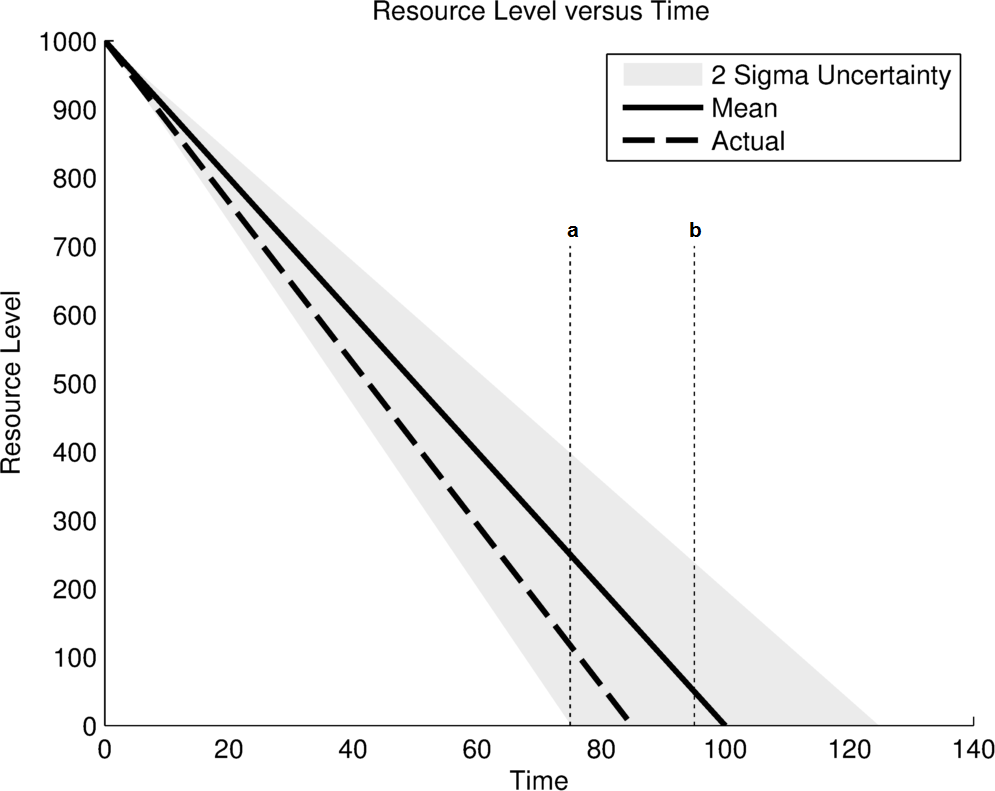}
  \caption{Level of a user agent showing predicted level, uncertainty and actual level. Replenishment at point $a$ is selected when using the stochastic cost estimate, while the deterministic cost estimate cannot differentiate between the two points. }
  \label{f:uncertainty}
\end{figure}

The stochastic forms of the objective functions show a clear advantage over the deterministic forms when considering the percentage of simulations in which none of the agents run out of fuel. As is shown in Fig.~\ref{sf:4_agent_2}, for a schedule length of 7 tasks, the stochastic forms of the objective functions were able to keep the agents from running empty in 80\% of the simulations, while the deterministic forms were only able to do so in 32.5\% and 25\% of the simulations for the DT and DR methods respectively. Where the stochastic forms of the objective functions choose low risk schedules, the deterministic forms have no way of differentiating between zero cost solutions. This is illustrated in Fig.~\ref{f:uncertainty}. The stochastic method is able to recognise point $a$ as lower risk than point $b$, and will therefore choose to replenish the user agent at point $a$ in preference to point $b$. However, a deterministic cost function treats them as equal as they both result in a zero cost. If the actual resource level is as shown and point $b$ is chosen, the user agent will exhaust the resource before it is replenished. 

In the 5-user agent scenario results, shown in Fig.~\ref{sf:5_agent}, the stochastic forms of the objective functions generally outperformed the deterministic forms in terms of both median and interquartile range for all of the schedule lengths. In addition, the DR and SR objective functions mostly outperformed the DT and ST objective functions respectively for the same horizon length. Similar to the 4-user agent scenario, the stochastic forms significantly outperformed the deterministic cost functions when considering the percentage of simulations with 100\% uptime, shown in Fig.~\ref{sf:5_agent_2}. 

In the 6-user agent scenario, shown in Fig.~\ref{sf:6_agent}, the DR and SR objective functions outperformed the DT and ST objective functions when using a planning horizon of 7 or 8 tasks. When 9 tasks are used, all of the objective functions provided comparable performance. The similar performance between the deterministic and stochastic forms of the objective functions is expected as there are too many user agents for the system to achieve 100\% uptime. Thus, the probability distributions for the time that each user agent is empty will be mostly in the positive domain, resulting in expected values that are close to the mean value of the distribution. As a result, the cost calculated by the stochastic forms of the objective functions should be similar to those calculated by the deterministic forms. 

From this study, it is clear that the stochastic forms of the objective functions produce superior results in the under-utilised and fully-utilised scenarios, where it is possible to prevent the user agents from running out of the resource. In addition, the ratio objective functions generally show better performance than the total tardiness objective functions when using shorter planning horizons. The difference in performance between the total tardiness and ratio objective functions decreases as the planning horizon is increased. However, using very long planning horizons is not feasible with A* due to the exponential growth of the search space, and other optimisation methods, such as branch and bound, should be used. 


\section{CONCLUSIONS} \label{s:conc}

This paper explored the optimisation of the replenishment agent's schedule for a SCAR scenario with a single replenishment agent. Deterministic and stochastic forms of the total weighted tardiness and ratio objective functions were used in conjunction with an A* search algorithm to find the optimal schedule for the replenishment agent. A computational study showed that the stochastic forms of the objective functions provided clear benefit in situations where the replenishment agent is under-utilised or fully-utilised. The performance of the objective functions was shown to generally increase as the number of tasks in the schedule was increased. The ratio objective functions showed superior performance to the total weighted tardiness objective functions when using shorter schedule horizons. The stochastic ratio objective function generally had the best performance across all scenarios, and is the recommended objective function for use in this form of the SCAR scenario. 

Possible areas of future research include developing more sophisticated heuristics to estimate the cost to complete a partial schedule, optimising scenarios with more user agents, and extending the optimisation process to a multi-replenishment agent system.



\bibliographystyle{IEEEtran}
\bibliography{IEEEabrv,ref}

\end{document}